\documentclass[letterpaper, 10 pt, conference]{IEEEtran}
\usepackage{stmaryrd}
\usepackage{amsfonts}
\usepackage{graphicx,times,psfig,amsmath} % Add all your packages here

\IEEEoverridecommandlockouts    % to create the author's affliation portion
\begin{document}

% paper title: Must keep \ \\ \LARGE\bf in it to leave enough margin.
\title{\ \\ \LARGE\bf  Neural Network Multitask Learning for Traffic Flow Forecasting
\thanks{Feng Jin and Shiliang Sun are with the Department
of Computer Science and Technology, East China Normal University,
Shanghai 200241, P. R. China
 (email: fjin07@gmail.com;slsun@cs.ecnu.edu.cn).}
\thanks {This work was supported in part by the
National Natural Science Foundation of China under Project 60703005,
and in part by Shanghai Educational Development Foundation under
Project 2007CG30.}}

\author{Feng Jin and Shiliang Sun}
% avoiding spaces at the end of the author lines is not a problem with
% conference papers because we don't use \thanks or \IEEEmembership
% use only for invited papers
%\specialpapernotice{(Invited Paper)}

% make the title area
\maketitle

\begin{abstract}
Traditional neural network approaches for traffic flow forecasting
are usually single task learning (STL) models, which do not take
advantage of the information provided by related tasks. In contrast
to STL,  multitask learning (MTL) has the potential to improve
generalization by transferring information in training signals of
extra tasks. In this paper, MTL based neural networks are used for
traffic flow forecasting. For neural network MTL, a backpropagation
(BP) network is constructed by incorporating traffic flows at
several contiguous time instants into an output layer. Nodes in the
output layer can be seen as outputs of different but closely related
STL tasks. Comprehensive experiments on urban vehicular traffic flow
data and comparisons with STL show that MTL in BP neural networks is
a promising and effective approach for traffic flow
forecasting.\end{abstract}

% no key words

\section{Introduction}
% no \PARstart

With the rapid development of modern economy, more and more people
start to use automobiles. Urban traffic congestion has become a
commonplace phenomenon, which brings a series of social problems to
our lives. If traffic conditions, especially the coming of peak
traffic flows, can be predicted accurately, people could respond in
advance to prevent roads from being jammed. Smooth and well-ordered
traffic will give a great convenience to the public and our society.
The establishment of a better traffic flow forecasting model is the
basis of predicting traffic flows to avoid the congestion situation.
For example, it could provide valuable traffic information to
Intelligent Transportation Systems (ITS) to anticipate congestion
occurrence as early as possible.

So far, people have raised a variety of methods for traffic flow
forecasting, such as nonparametric methods [1], local regression
models [2,3], neural network approaches [4], fuzzy-neural approaches
[5],  Markov chain models [6], and Bayesian network approaches [7].
In this paper, exploring the performance of neural networks from a
new point of view for traffic flow forecasting is our concern.

A neural network (NN) is an approximation and variation to a
biological neural system but is highly simplified. It has various
intelligent processing functions such as learning, memorizing and
predicting.
 NNs can solve modeling problems for complicated systems which are uncertain and
seriously non-linear [8]. The traditional neural network approach
for traffic flow forecasting is to learn a task at a time [9]. It is
a single task learning (STL) model which neglects the potential and
rich information resources hidden in other related tasks. The
opposite is the multitask learning (MTL) neural network approach
which has more than one output [10]. In MTL, the task considered
most is called the main task, while others are called extra tasks.
MTL can improve generalization performance of neural networks by
integrating some field-specific training information contained in
the extra tasks [11]. In this paper, we focus on using MTL
backpropagation (BP) networks to carry out traffic flow modeling and
forecasting. Experiments with encouraging results show that this
approach is considerably effective for traffic flow modeling and
forecasting.

The rest of this paper is organized as follows. MTL and its benefits
for traffic flow forecasting are introduced in Section II. Then we
give the model construction mechanism, and report experimental
results in Section III. Section IV summarizes this paper and gives
future research directions.

\section{MTL}

\subsection{MTL and NN}

MTL is a form of inductive transfer whose main goal is to improve
generalization performance [12]. It uses the domain-specific
information which is included in the training signals of extra tasks
to improve generalization. In fact, the training signals for the
extra tasks serve as an inductive bias [13]. In other words, that
bias is used to improve the generalization accuracy in order to
perfectly complete the main task. It is because that the helping
information which is provided by inductive bias  is stronger than
the one gained without the extra knowledge. As reported, better
generalization can often be yielded by employing MTL if there is
only a fixed training set [14]. MTL also can be used to reduce the
number of training patterns needed to achieve some fixed level of
performance [15].

Normally, most learning methods such as traditional neural networks
only have one task. When we want to solve a complicated problem, it
could be split into a number of small, appropriately independent
subproblems to learn [16]. This may ignore a potentially rich source
of information contained in the training signals of other tasks
drawn from the same domain [17]. In fact, it is believed that
most-real-world problems are multitask problems and performances are
being sacrificed when we treat them as single problems. Therefore,
we introduce MTL NN which has more than one output to predict
traffic flows. In a MTL NN, all tasks are trained in parallel using
a shared representation. And the information contained in these
extra training signals can help the hidden layer learn a better
internal representation for the main task. MTL has the potential to
meliorate the prediction accuracy of the main task and improve the
generalization of the entire network through learning the extra
tasks.

With the inductive bias provided by the extra tasks, MTL is
applicable to any learning methods to improve generalization
performance. Besides, the speed of learning, and the intelligibility
of learned models are also ameliorated via employing MTL. This paper
concentrates on improving generalization accuracy of neural networks
using the inductive transfer paradigm MTL.

\subsection{MTL for traffic flow forecasting}
MTL is broadly applicable. One of the applications is using the
future to predict the present [18]. Time series prediction is the
subclass of using the future to predict the present. MTL used for
series prediction is a new method for traffic flow forecasting whose
core idea is to make prediction for the same task at different time.
Using
 MTL NN to make time series prediction, the simplest approach is
to use a network with a number of outputs, each output corresponding
to one task which is at different time. In addition, the output used
for prediction can be the middle one so that there are tasks earlier
and later than it trained in the net [11].

The flow of one traffic junction at any time has a correlation with
the flow of its contiguous moment. Therefore, using time series
prediction to forecast traffic flow has significance. We can
arbitrarily choose the flow of m continuous time instants to
forecast their next moment's flow. The way we use MTL is that we
predict the vehicle flow at time n (denoted by t(n)) by selecting
t(n-1) and t(n+1) as well as outputs. Therefore, forecasting t(n) is
treated as the main task in the network. We generally select the
extra tasks which definitely have connections with the main tasks.
Here, we consider that t(n-1) and  t(n+1) have certain relationship
with the main task and choose them as extra tasks. They play an
inductive bias role for the main task so as to increase the
forecasting accuracy of t(n).

\section{ Model construction and experiments}
\subsection{The source of data}

Now, we use the idea of time series prediction for traffic flow
forecasting, and the net is BP neural network using MTL. It means
that we forecast future traffic flow at given road links from the
historical data of themselves. The difference from ordinary neural
network is that the current uses a method of MTL. It has more than
one output.

The data used for analysis are the vehicle flow rates of discrete
time series which were recorded every 15 min. Data's unit is
vehicles per hour (veh/h). We choose a fraction taken from one urban
traffic map of highways to verify the approach of MTL. The fraction
is shown in Fig.~1. Each circle node in the sketch map denotes a
road junction. An arrow shows the direction of traffic flow, which
reaches the corresponding road link from its upstream link. Paths
without arrows are of no traffic flow records. The raw data are from
March 1 to March 31, 2002, totaling 31 days [7]. Considering the
malfunction of detector or transmitter, we screened the days with
empty data in view of evaluation. The remaining data for use are of
25 days and totaling 2400 sample points. We select the first 2112
samples points and treat them as training data. The rest are used
for test data.

\begin{figure}[h]
  % Requires \usepackage{graphicx}
  \includegraphics[width=3.35in]{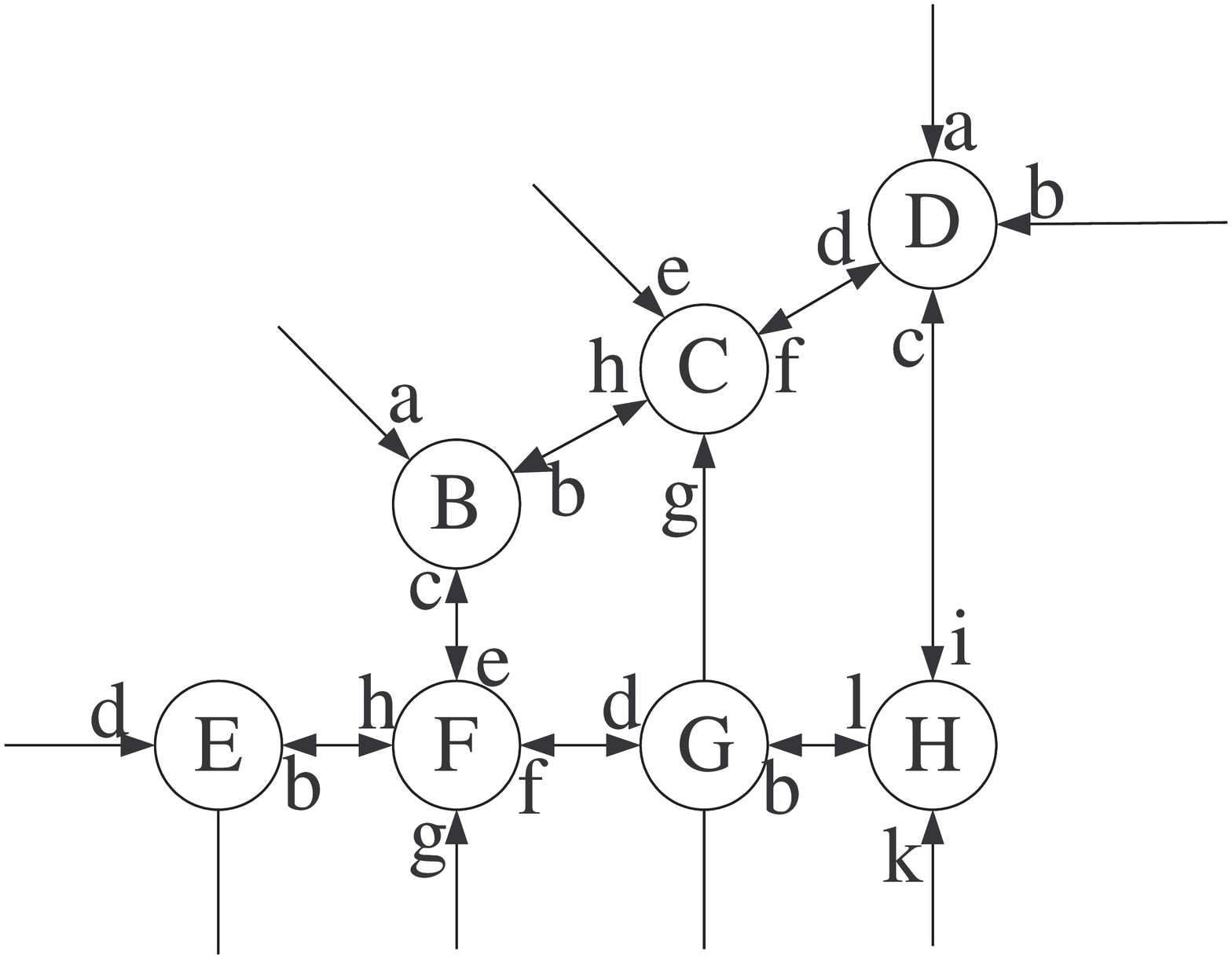}\\
  \caption{Patch taken from the East Section of the Third Circle of Beijing City Map where UTC/SCOOT systems are placed. For convenience, roads and flows are coded with English characters.}\label{fig_sim}
\end{figure}

\subsection{Model building}
The model building of the traffic flow forecasting can be concluded
as follows:

(1). A three-layer NN is chosen. The reason is that it can
approximate to any continuous function, as long as the appropriate
number of hidden layer neurons and right activation functions are
used. If the network layers increase, the network will become
complicated and the training time will increase. Therefore, a
network which has an input layer, a hidden layer and an output layer
is selected.

(2). Traffic flows of five seriate time instants are adopted as
inputs. A function which can make data normalize to deal with the
sample points is applied. After doing that, the original data are
normalized to the range -1 from 1. At the same time, the pace of
training may accelerate. After normalization, the initial weights of
network will not be big, enabling network performance and the
ability of generalization better. For the hidden layer we select
fifteen neurons according to practical training performance.

(3). Sigmoid function is selected as the specific activation
function between input layer and hidden layer. The form of sigmoid
function is described as
\begin{equation}
\ f(x)=\frac{2}{1+e^{-2x}}-1.
 \label{eq:eq2}
\end{equation}
This function can make neuronal inputs map to a range from -1 to 1 .
Since it is a differentiable function which is suitable to train NN
using the algorithm of BP. The activation function between hidden
layer and output layer is a linear function whose form is shown as

\begin{equation}
\ f(x)=x.
 \label{eq:eq2}
\end{equation}
It is also a differentiable function. We can get arbitrary value
from the function as its outputs.

(4). Levenberg-Marquardt algorithm is selected to train the network
[19,20]. Because it has comparatively fast convergence speed and
high precision which can rectify network's weights and threshold
better. The adjustment formula is shown as

\begin{equation}
\ x_{k+1}=x_{k}-[J^TJ+\mu{I}]^{-1}J^Te,
 \label{eq:eq2}
\end{equation}where J is
a Jacobian matrix whose elements are the network error's first
derivatives with respect to weights and the corresponding threshold,
e is the network error vector, and $ \mu$ is a scalar quantity which
is initialized.

 There are many parameters in the training function including the
 largest training epoch, training
time, and network error target which can be all acted as the
stopping conditions of training. If training time of the network has
no strict demands, the largest training epochs which can make the
network converge are selected as the only guideline of the training
stopping by observing the changes in network training error. Since
the establishment of the preferable network model is based on the
results of a large number of practical trainings, so only when we
change a parameter to inspect the effect which is brought to the
network, can we build a better network model [21].

(5). The constructed MTL network model is given in Fig.~2. The
network has three outputs and the forecasting of t(n) is severed as
main task. We employed STL as a comparative method and the STL
network model is given in Fig.~3. As is shown, the net only has one
task.The difference between that two NNs is only in their output
layer. STL NN has one neuron as its output, while MTL NN not only
has main task, but also has extra task.
\begin{figure}[h]
  % Requires \usepackage{graphicx}
  \includegraphics[width=3.35in]{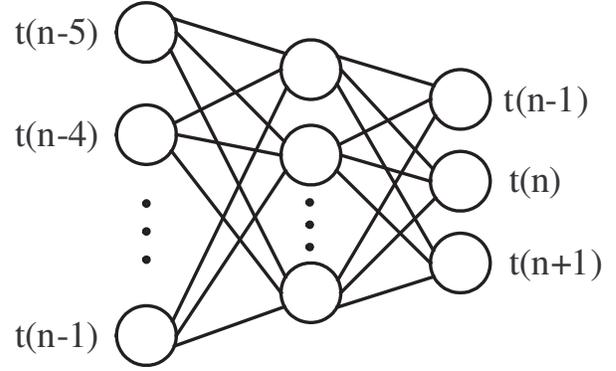}\\
  \caption{NN using MTL to forecast t(n).}\label{fig_sim}
\end{figure}

\begin{figure}[h]
  % Requires \usepackage{graphicx}
  \includegraphics[width=3.35in]{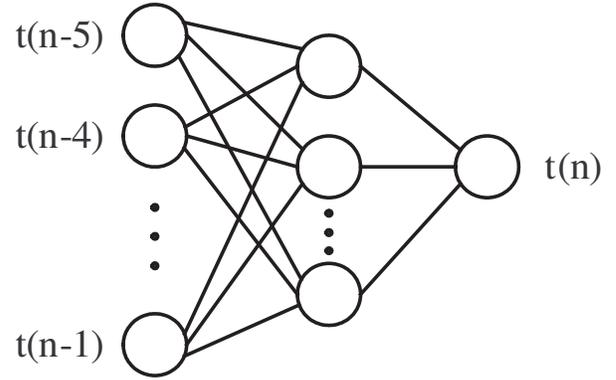}\\
  \caption{ NN using STL to forecast t(n).}\label{fig_sim}
\end{figure}

\subsection {Experimental results}
We take vehicle flow data $\mathbf{B_b}$  as an instance to show our
modeling mechanism.  $\mathbf{B_b}$   represents the vehicle flow
from upstream junction C to downstream junction B. The traffic flow
figure of all sample points in 25 days is depicted in Fig.~4.

The training set and test set have been described, we predict
traffic flow only for single inflow of a junction. The network is
trained by using the first 22 days' data to forecast the traffic
flow of later 3 days. Experiments are firstly done by using MTL.
According to practical training, setting training parameters of the
NN. The goal of training error is initially defined as 0.006. In
course of the experiment, training error of the network is shown in
Fig.~5.

\begin{figure*}
  % Requires \usepackage{graphicx}
  \includegraphics[width=6.90in]{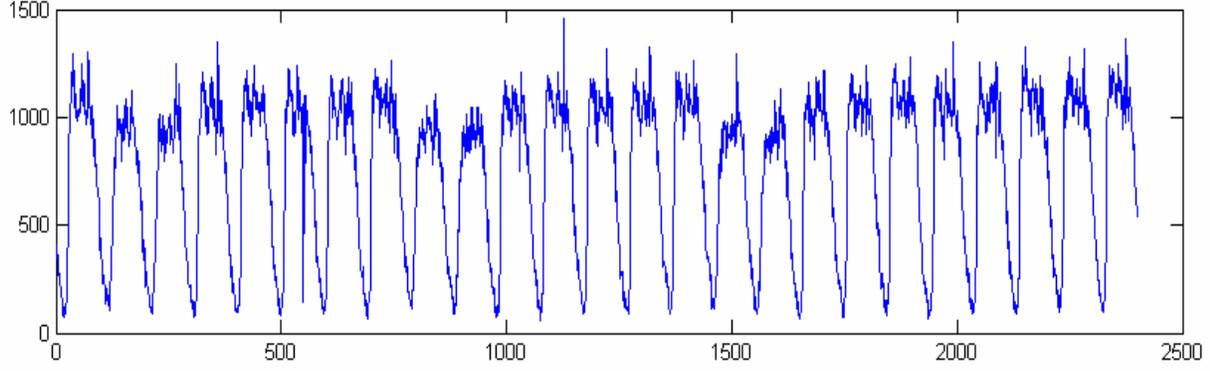}\\
  \caption{ The original value of traffic flow $\mathbf{B_b}$.}\label{fig_sim}
\end{figure*}

\begin{figure}[h]
 % Requires \usepackage{graphicx}
  \includegraphics[width=3.80in]{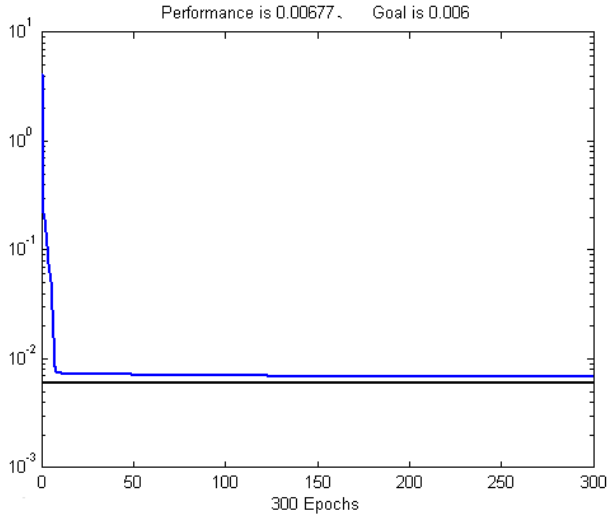}\\
  \caption{  Network training error of $\mathbf{B_b}$.
  The  curve represents the changed condition of error in the course of training,
   the horizontal line denotes the goal of training error.}\label{fig_sim}
\end{figure}

From Fig.~5, we can see that network error hardly changes after 125
training steps. In order to increase the training accuracy, 300
epochs are selected as the maximal training epochs in the end. It
also means that the training will stop after experiencing 300
training epochs.

The final forecasting result of the MTL network  for the last 3
days' data of link $\mathbf{B_b}$ is shown in Fig.~6.

\begin{figure}[h]
  % Requires \usepackage{graphicx}
  \includegraphics[width=3.80in]{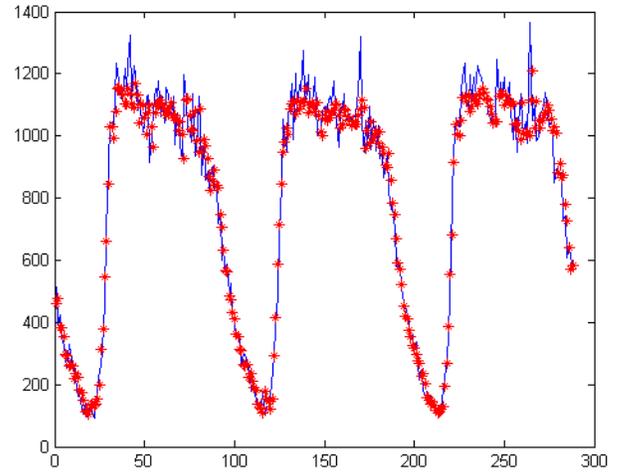}\\
  \caption{ MTL traffic flow forecasting result for the last 3 days' data of road link $\mathbf{B_b}$.}\label{fig_sim}
\end{figure}

Root mean square error (RMSE) is used to evaluate prediction
performance of the networks. The expression is shown as

\begin{equation}
\ RMSE=({\sum_{i=0}^n}{\frac{(t(i)-\alpha(i))^2}{n}})^{1/2},
 \label{eq:eq2}
\end{equation} where
$\alpha$ is the estimate of n-dimensional vector  t.
  The performance measure RMSE of $\mathbf{B_b}$  in MTL  can be calculated as the following:

\begin{equation}
\ RMSE_M=70.25.
 \label{eq:eq2}
\end{equation}

Now, we employed STL as a comparative method. The final forecasting
result of the STL network  for the last 3 days' data of link
$\mathbf{B_b}$  is shown in Fig. 7. Similarly the RMSE is:

\begin{equation}
\ RMSE_S=77.46.
 \label{eq:eq2}
\end{equation}

\begin{figure}[h]
  % Requires \usepackage{graphicx}
  \includegraphics[width=3.80in]{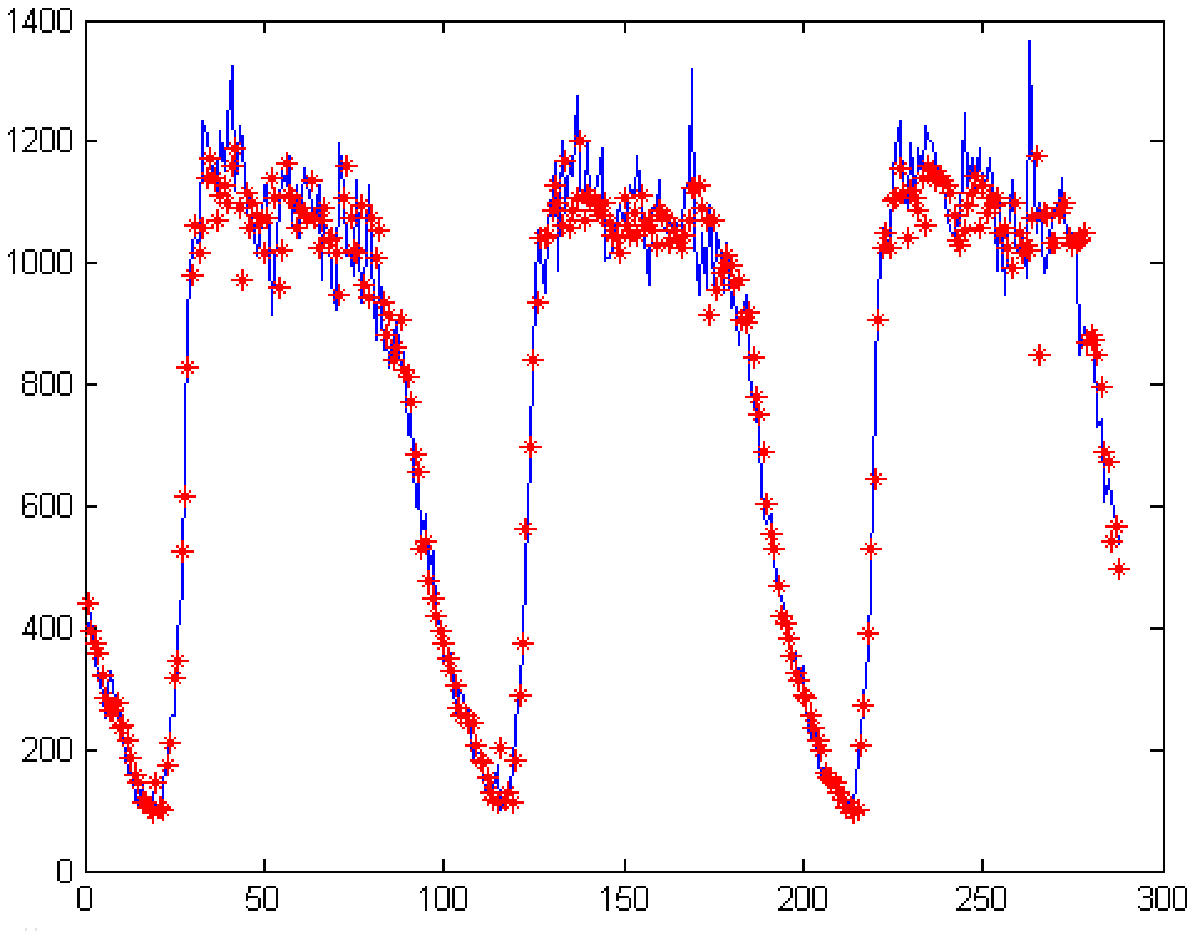}\\
  \caption{  STL traffic flow forecasting result for the last 3 days' data of road link $\mathbf{B_b}$.}\label{fig_sim}
\end{figure}

Known from formula (5) and formula (6), MTL is better than STL in
forecast accuracy.
 Furthermore, the performance improvement of MTL is defined as follows:

\begin{equation}
\ {{e}_B}_b=\frac{{RMSE}_S-{RMSE}_M}{{RMSE}_S} = 9.31\%
 \label{eq:eq2}
\end{equation}
 This is meant that MTL is better than STL in prediction accuracy by 9.31 percent.
In order to objectively evaluate the performance of STL and MTL for
traffic flow forecasting, we also conduct experiments on other road
links. Likewise, we select the largest training epochs which can
make the network converge as the only guideline of the training
stopping. The other parameters are all based on the results of a
large number of practical trainings.  The results are given in Table
I.

\begin{table}[h]
\caption{Comparison results  with STL} %\cite{cit:mic77}
 %(40unit system with valve-point effects)}
\begin{center}
\begin{tabular}{|c|c|c|c|c|c|c|}
\hline \multicolumn{1}{|c|}{\!RMSE\!} &
\multicolumn{1}{|c|}{$\mathbf{C_f}$}
&\multicolumn{1}{|c|}{$\mathbf{D_b}$}
&\multicolumn{1}{|c|}{$\mathbf{F_f}$}
&\multicolumn{1}{|c|}{$\mathbf{G_b}$}
&\multicolumn{1}{|c|}{$\mathbf{H_i}$}
&\multicolumn{1}{|c|}{$\mathbf{B_b}$}\\
\hline
%&$\mathbf{C_f}$&$\mathbf{D_b}$&$\mathbf{F_f}$&$\mathbf{G_b}$&$\mathbf{H_i}$&$\mathbf{B_b}$\\\hline

STL   &  $93.03$  &  $55.04$  &  $85.28$ & $ 84.89$ & $92.28$ &
$77.46$ \\\hline
 MTL   &  $86.26$  &  $52.85$  &$82.02$ & $82.90$&
$88.02$ & $70.25$ \\\hline
 e  &  $7.28$\%$$  &  $3.98$\%$$  &  $3.82$\%$$ & $2.34$\%$$ &
$4.62$\%$$ & $9.31$\%$$ \\ \hline
 %MFEP  &  $1056.8$  &  $1054.2$  &  $123489.7$ & $124356.5$ & $122647.6$ \\ \hline
 %IFEP  &  $632.67$  &  $630.36$  &  $123382.0$ & $125740.6$ & $122624.4$ \\ \hline
 %TM    &  $94.28$  &  $91.16$  &  $123078.2$ & $124693.8$ & $122477.8$ \\ \hline
\end{tabular}
%\label{tab-liu2}
\end{center}
\end{table}

 From the results, we can see
that RMSE of the forecasting results all diminish after using MTL.
It implies that the forecasting precision is improved. Therefore,
MTL NN can be used for traffic flow forecasting to get relatively
precise results. It is useful for ITS to deal with the problem of
traffic congestion.

\section{Conclusion}
Traditional forecasting methods are generally to anticipate with
STL. They fail to attach importance to a potential, rich and
available information resources which can be obtained from training
information of other tasks in the same area. Usually, those results
are not ideal. MTL mentioned in this paper is synchronal to train
more than one task and can take full advantage of training
information in the extra tasks to improve the generalization of the
network. It makes the net have higher forecasting accuracy.

From above experiments, we can see that the forecasted results of
traffic flows  are closer to the true value when multitask learning
is used in the network. It illuminates that multi-task learning for
time series prediction of traffic flow is very practical. In the
future, comparing the neural network MTL approach with other
methods, such as kernel regression [1,11], Bayesian networks [7]
will be investigated. Besides, incorporating information from
neighbor road links would also be studied [22].

%\section*{Acknowledgment}
 %The author would like to thank Dr. sun for his
%sincere help.Thanks also go to the anonymous reviewers for their
%valuable advices.

%\bibitem{Siedited} J. Si, A. Barto, W. Powell, and D. Wunsch, Editors,
 %   {\it Handbook of Learning and Approximate Dynamic Programming},
  %      New York: IEEE and Wiley, 2004.

%\bibitem{cit:wer9213} P. J. Werbos, ``Approximate dynamic programming for
 %   real-time control  and neural modeling,'' in
  %      \handb (Chapter~13), Edited by D.~A.~White
   % and D.~A.~Sofge, New York, NY: Van Nostrand Reinhold, 1992.

%``Multiple model fault tolerant control using globalized dual
%heuristic programming,''

% that's all folks
\end{document}